\documentclass[10pt,twocolumn,letterpaper]{article}

\usepackage{iccv}              
\usepackage{float}
\usepackage{stfloats}
\usepackage{soul}
\usepackage{makecell} 
\usepackage{amsmath}
\usepackage{amssymb}
\usepackage{multirow}
\usepackage{colortbl}
\usepackage{graphicx}
\usepackage{pifont}
\usepackage{tikz}        
\usepackage{pgfplots}    
\pgfplotsset{compat=1.18} 

\definecolor{iccvblue}{rgb}{0.21,0.49,0.74}
\usepackage[pagebackref,breaklinks,colorlinks,allcolors=iccvblue]{hyperref}

\def\paperID{2749} 
\def\confName{ICCV}
\def\confYear{2025}

\title{UST-SSM: Unified Spatio-Temporal State Space Models for Point Cloud Video Modeling}

\author{
    \small
    Peiming Li\textsuperscript{1}\quad
    Ziyi Wang\textsuperscript{1}\quad 
    Yulin Yuan\textsuperscript{2}\quad 
    Hong Liu\textsuperscript{1}\quad 
    Xiangming Meng\textsuperscript{2}\quad
    Junsong Yuan\textsuperscript{3}\quad 
    Mengyuan Liu\textsuperscript{1$\dagger$}\\
    \small
    $^1$State Key Laboratory of General Artificial Intelligence, Peking University, Shenzhen Graduate School\\
    \small
    $^2$The Zhejiang University-University of Illinois Urbana-Champaign Institute, Zhejiang University\quad \\
    \small
    $^3$State University of New York at Buffalo \\
    {\tt \small
    \{lipeiming1001,ziyiwang\}@stu.pku.edu.cn\quad 
    yulinyuan@zju.edu.cn\quad
    hongliu@pku.edu.cn\quad 
    }\\
    {\tt \small
    xiangmingmeng@intl.zju.edu.cn\quad
    jsyuan@buffalo.edu\quad  
    liumengyuan@pku.edu.cn\quad  
    }
}

\begin{document}
\maketitle

\def\thefootnote{}\footnotetext{$\dag$ Corresponding author: Mengyuan Liu (liumengyuan@pku.edu.cn)} 

\begin{abstract}
Point cloud videos capture dynamic 3D motion while reducing the effects of lighting and viewpoint variations, making them highly effective for recognizing subtle and continuous human actions. Although Selective State Space Models (SSMs) have shown good performance in sequence modeling with linear complexity, the spatio-temporal disorder of point cloud videos hinders their unidirectional modeling when directly unfolding the point cloud video into a 1D sequence through temporally sequential scanning. To address this challenge, we propose the Unified Spatio-Temporal State Space Model (UST-SSM), which extends the latest advancements in SSMs to point cloud videos. Specifically, we introduce Spatial-Temporal Selection Scanning (STSS), which reorganizes unordered points into semantic-aware sequences through prompt-guided clustering, thereby enabling the effective utilization of points that are spatially and temporally distant yet similar within the sequence. For missing 4D geometric and motion details, Spatio-Temporal Structure Aggregation (STSA) aggregates spatio-temporal features and compensates. To improve temporal interaction within the sampled sequence, Temporal Interaction Sampling (TIS) enhances fine-grained temporal dependencies through non-anchor frame utilization and expanded receptive fields. Experimental results on the MSR-Action3D, NTU RGB+D, and Synthia 4D datasets validate the effectiveness of our method. Our code is available at \url{https://github.com/wangzy01/UST-SSM}.
\end{abstract}

    
\section{Introduction}



\begin{figure}[!thbp]
	\centering 
	\begin{tabular}{c}		
		\includegraphics[width=7.5cm]{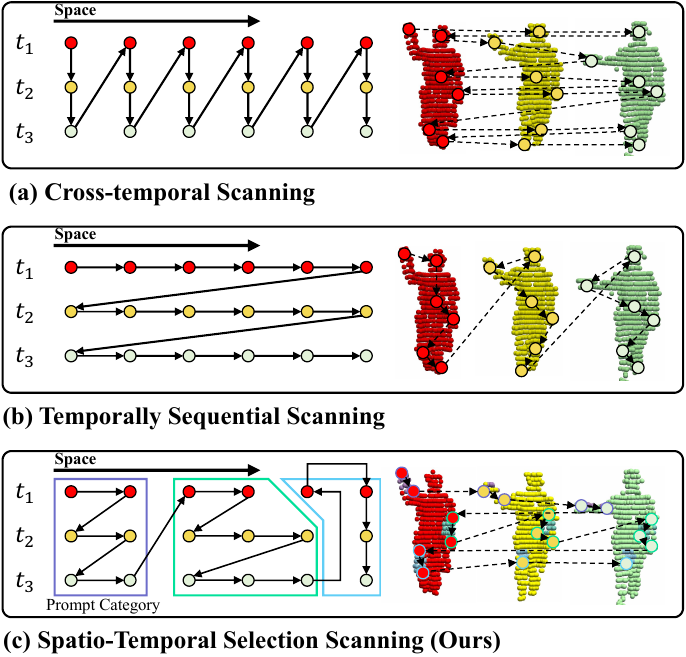}\\
	\end{tabular}%
    \vspace{-2mm}
	\caption{Illustration of Spatio-Temporal Selection Scanning and comparison with previous scanning strategies \cite{mamba4d,Mamba3D,pointmamba}. \textbf{(a) Cross-temporal Scanning:} Points across the frame are scanned according to coordinates. \textbf{(b) Temporally Sequential Scanning:} Points are first scanned within each frame, then along the temporal dimension. \textbf{(c)  Spatio-Temporal Selection Scanning (Ours):} The scanning first occurs within semantic-aware sequences through prompt-guided clustering. Within each cluster, Hilbert sorting is performed to preserve local geometry, followed by the temporal dimension.}\label{fig:fig1}%
    \vspace{-4mm}
\end{figure}%

\begin{figure}[!thbp]
	\centering 
	\begin{tabular}{c}		
		\includegraphics[width=7.5cm]{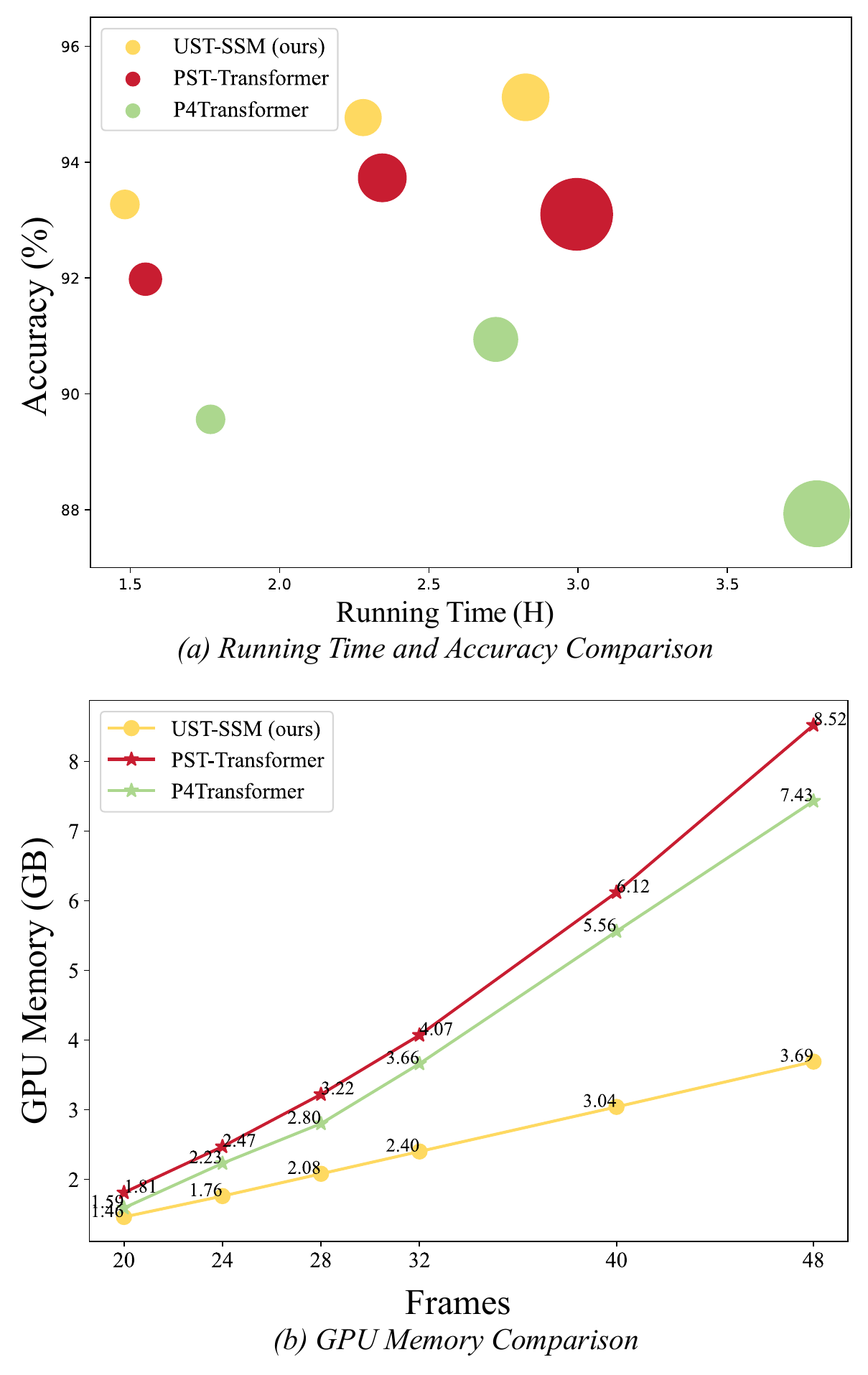}\\
	\end{tabular}%
        \vspace{-2mm}
	\caption{Comparison of UST-SSM with P4Transformer and PST-Transformer. \textbf{(a) Comparison on running time and accuracy (the circle size represents the GPU memory) and (b) comparison on GPU memory:} Our UST-SSM, with linear complexity, achieves optimal performance in terms of runtime, memory usage, and accuracy, as detailed in \cref{sec:3daction}.}\label{fig:fig2}%
    \vspace{-6mm}
\end{figure}%

\label{sec:intro}
Recent advances in depth sensing technologies have driven growing interest in point cloud videos analysis for robotics \cite{101,102,103} and autonomous systems \cite{104,105,106}. Point cloud videos inherently resist lighting/viewpoint variations while preserving 3D motion dynamics, making them particularly suitable for human action recognition.

Recent advances in point cloud video modeling \cite{Wei_2022_WACV, pptr, psttrans, p4trans, MeteorNet, Fan_Yu_Yang_Kankanhalli_2022, scannet} leverage CNNs or Transformers. However, CNNs \cite{MeteorNet} struggle with modeling long-term dependencies, and P4Transformer \cite{p4trans} and PST-Transformer \cite{psttrans} consume significant memory for long sequences. Recently, selective state space models (SSMs) have shown competitive performance compared with Transformers in NLP, offering linear complexity and significantly reduced memory usage. SSM-based methods \cite{Mamba3D, pointmamba} have successfully applied SSMs to 3D point cloud analysis by serializing the spatial structure, but they focus solely on the 3D spatial dimension and neglect the temporal aspect. Mamba4D \cite{mamba4d} replaces the classical transformer by employing a predefined temporally sequential scanning, unfolding the point cloud videos into 1D sequence. However, each point in Mamba is modeled based solely on its preceding point in the scanning sequence, which causes temporally sequential scanning to restrict the interaction range of query points to information from previous frames. Furthermore, previous research \cite{mambairv2} has shown that Mamba experiences long-range attenuation in token interactions, meaning that interactions between tokens that are spatially and temporally distant are reduced. As a result, even distant but relevant points from earlier scans cannot be effectively utilized by the query points. In contrast, our UST-SSM employs Spatial-Temporal Selection Scanning, which clusters distant yet similar spatio-temporal points, overcoming the limitations of temporal sequential scanning while mitigating the long-range attenuation of similar points. Fig.\ref{fig:fig1} highlights the key differences between previous scanning strategies and STSS. 
As demonstrated in \cref{fig:fig2}, our method outperforms Transformer-based approaches in both performance and efficiency.

However, effectively modeling the spatio-temporal characteristics of point cloud videos presents unique challenges: (1) Non-unified spatio-temporal disorder: The absence of consistent spatio-temporal ordering in point cloud videos fundamentally conflicts with the unidirectional scanning paradigm of SSMs, as existing scanning strategies fail to achieve the desired spatio-temporal order. (2) Local geometry loss: Existing SSM-based methods' serialization strategies discard crucial geometric relationships between neighboring points. (3) Temporal interaction limitation: Traditional temporal sampling strategies create fragmented temporal contexts that hinder motion analysis. 

Our proposed UST-SSM unifies spatio-temporal modeling for point cloud videos and addresses these issues through three key technical contributions: (1) Spatio-Temporal Selection Scanning (STSS): Point cloud videos lack consistent temporal alignment, and spatial disorder complicates temporal modeling. When extending 3D point cloud methods to 4D, previous methods typically perform spatial sorting followed by temporally sequential scanning. However, points that differ in temporal dimensions may actually represent the same point or exhibit higher similarity. After sorting, these points are placed far apart, leading to the problem of long-range attenuation of similar points in unidirectional modeling. STSS introduces a lightweight prompt network to cluster points based on semantic similarity rather than temporal proximity. This enables coherent sequence formation by grouping spatio-temporally distant but semantically related points, while preserving local geometry through Hilbert-curve spatial ordering within clusters. (2) Spatio-Temporal Structure Aggregation (STSA): The unidirectional sorting in SSMs relies solely on preceding points in the scan sequence, neglecting local geometric details. STSA actively recovers geometric details through spatio-temporal KNN neighborhood feature propagation, compensating for information loss during serialization. (3) Temporal Interaction Sampling (TIS): Traditional temporal sampling strategies usually ignore the fine-grained time dependence between frames, which limits the time interaction between frames. TIS innovatively leverages non-anchor frames through multi-stride sampling, expands the temporal receptive field, and improves temporal interaction within the sampled sequence. This spatio-temporal structure extends SSMs' sequential modeling capabilities to point cloud videos analysis, excelling in tasks that require long-term structure capture.


We evaluate our UST-SSM on the MSR-Action3D \cite{msr}, NTU RGB+D \cite{ntu60}, and Synthia 4D \cite{Minkowski} datasets. Experimental results demonstrate the effectiveness of our approach. The contributions of this paper are threefold:
\begin{itemize}
\item We propose UST-SSM, an SSM-based model designed to unify spatio-temporal modeling for point cloud videos, in order to achieve a spatio-temporal ordering suitable for SSMs. Spatio-Temporal Selection Scanning (STSS) organizes unordered point cloud videos by clustering and enhancing interactions among spatio-temporally distant yet spatially similar points.
\item We introduce Spatio-Temporal Structure Aggregation (STSA), which compensates for the spatial features overlooked during sorting by incorporating local geometric details. Temporal Interaction Sampling (TIS) extends the time receptive field and improves temporal information exchange and fine-grained feature capture.
\item Extensive experiments demonstrate that the proposed UST-SSM extends SSMs' sequential modeling capabilities to point cloud videos analysis, enhancing computational efficiency and reducing resource consumption, particularly in long-term spatio-temporal modeling.
\end{itemize}

\vspace{-2mm}
\section{Related Work}
\textbf{Point Cloud Video Modeling.} Early approaches convert point cloud videos into 3D voxel sequences \cite{Fast,Minkowski}, though voxelization often leads to a loss of geometric detail. Most existing methods rely on convolutional networks or vision transformers for direct point cloud video modeling. MeteorNet \cite{MeteorNet} was the first to extend PointNet++ to the temporal dimension, allowing it to handle point cloud videos. To increase the receptive field, P4Transformer \cite{p4trans} leverages transformers to avoid explicit point tracking but does not encode the full spatio-temporal structure. PST-Transformer \cite{psttrans} addresses this by encoding features based on the spatio-temporal displacement between reference points and all points in the video. PPTr \cite{pptr} introduces geometric planes as mid-level representations to improve efficiency, enhancing performance for dynamic scenes and longer video sequences. However, it does not mitigate the quadratic complexity of transformers. In this work, we propose a novel point cloud video backbone based on SSMs, addressing challenges in computational efficiency and long-term spatio-temporal modeling.\\
\textbf{State Space Models.} State Space Models (SSMs) \cite{Gu2021EfficientlyML,NEURIPS2021_05546b0e} have shown great promise in sequence modeling tasks. Early models like HiPPO \cite{HIPPO} and LSSL \cite{NEURIPS2021_05546b0e} established foundational techniques for SSMs through projection and optimization methods. Structured State Space Sequence Models (S4) \cite{Gu2021EfficientlyML} broadened the applicability of SSMs in fields like speech and language, allowing SSMs to compete with transformers in long-sequence tasks. Advances such as S4D and S5 have introduced diagonal matrices and MIMO structures, making SSMs more efficient for parallel processing. Mamba \cite{mamba} significantly enhances SSM efficiency in long-sequence modeling through selective state space mechanisms and hardware-aware optimizations. This advancement has inspired various applications in the visual domain, including image processing \cite{visionmamba,vmamba}, video understanding \cite{videomamba,videomambasuitestate}, generative models \cite{zigma}, and 3D point clouds \cite{Mamba3D,pointmamba}. However, research on applying SSMs to point cloud videos remains limited. Mamba4D \cite{mamba4d} directly applies Mamba on top of P4Transformer. However, each point is modeled based solely on its preceding point in the scanning sequence, which restricts the interaction range of query points to information from the previous frame due to the temporally sequential scanning. In this work, we further investigate the potential of SSMs for point cloud video analysis and introduce Spatial-Temporal Selection Scanning, a method that clusters distant yet spatially similar points across time, effectively overcoming the constraints of sequential temporal scanning.

\section{Method}

\subsection{Overview}
\label{sec:overview}
\begin{figure*}[htbp]
	\centering 
	\begin{tabular}{c}		
		\includegraphics[width=17cm]{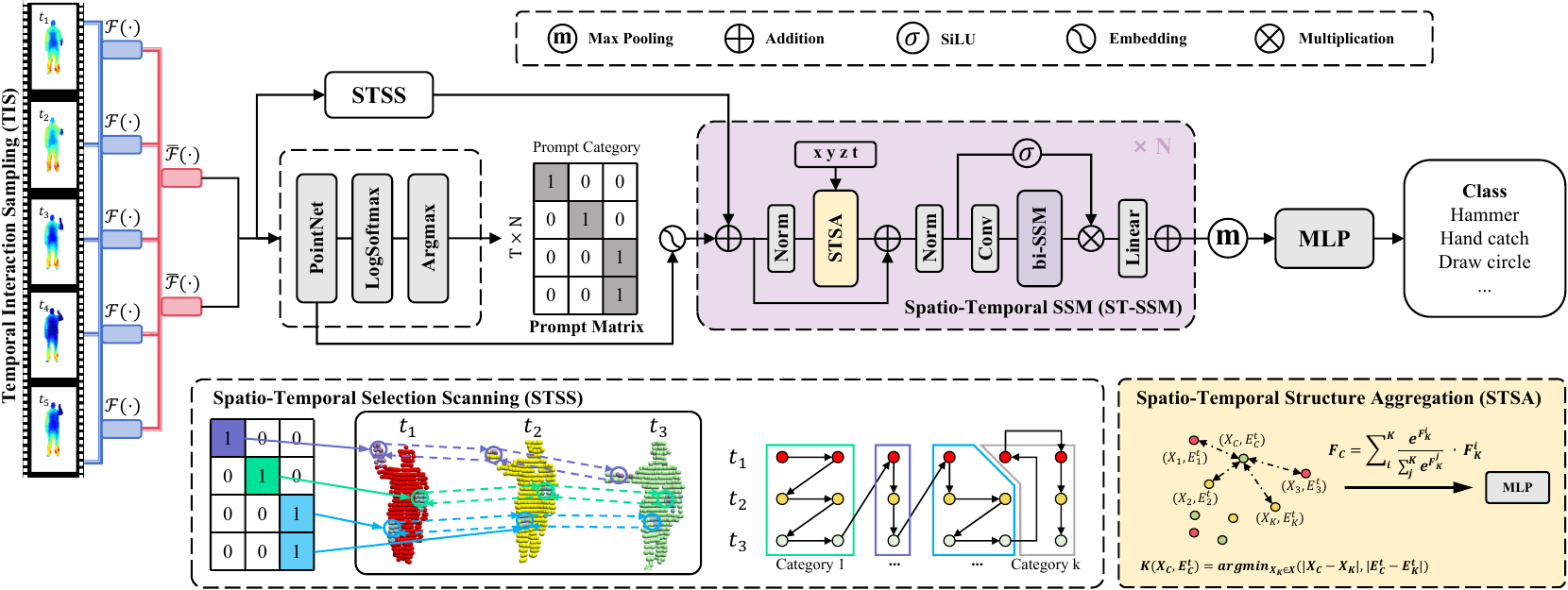}\\
	\end{tabular}%
	\caption{\textbf{Overview of UST-SSM:} First, Temporal Interaction Sampling is applied to the point cloud sequence. The downsampled sequence is then serialized using Spatio-Temporal Selection Scanning, followed by processing with ST-SSM and Spatio-Temporal Structure Aggregation. Finally, a Multi-Layer Perceptron is used for classification and prediction. }\label{fig:framework}%
    \vspace{-2mm}
\end{figure*}

Traditional SSMs face three fundamental challenges in point cloud video modeling: (1) \textit{Non-unified spatio-temporal disorder} from unaligned dynamic points disrupts unidirectional scanning. (2) \textit{Local geometry loss} caused by serialization compromises fine-grained details. (3) \textit{Fragmented temporal context} in conventional sampling limits motion understanding. Our UST-SSM addresses these through synergistic components:
\begin{itemize}
    \item \textbf{Temporal Interaction Sampling (\cref{sec:3.1.3})}: Solves Challenge (3) by constructing continuous temporal contexts through multi-stride sampling.
    \item \textbf{Spatio-Temporal Selection Scanning (\cref{sec:3.1.1})}: Addresses Challenge (1) via prompt-guided clustering.
    \item \textbf{Spatio-Temporal Structure Aggregation (\cref{sec:3.1.2})}: Mitigates Challenge (2) through 4D KNN and spatio-temporal feature propagation.
\end{itemize}
The overall process of UST-SSM is shown in \cref{fig:framework}.

\subsection{Temporal Interaction Sampling}
\label{sec:3.1.3}
Due to the unordered variations of high-frequency and low-frequency information in the temporal dimension of video sequences, traditional single-stride sampling strategies with small receptive fields tend to overlook fine-grained temporal dependencies between frames and fail to fully capture temporal context. This limitation results in insufficient interaction between frames. Temporal Interaction Sampling (TIS) addresses these issues through non-anchor frame utilization and expanded receptive fields, enabling better temporal interaction across the sampled sequence. 

Given a point cloud sequence \( \mathbf{X} = \{T_1, T_2, \dots, T_n\} \), single-stride sampling (\eg stride=2) in previous methods creates temporal islands where anchor frames \( \{T_2, T_4,...\} \) only interact with immediate neighbors. This restrictive temporal window fails to exploit the full range of temporal dependencies between frames, formulated as:
\begin{equation}
\begin{aligned}
\mathbf{Feat}_{T_{2i}} &=  \mathcal{F}(\mathcal{S}(T_{2i-1},T_{2i},T_{2i+1})),
\end{aligned}
\end{equation}
where \( i \in \{1, 2, \dots, n/2\} \), \( \mathcal{S} \) and \( \mathcal{F} \) represents the spatial sampling and feature extraction operations. To enhance the utilization of non-anchor frames and expand the temporal receptive field, TIS first performs temporal sampling with a stride of 1, which aggregates information from consecutive frames. This can be formulated mathematically as:
\begin{equation}
\begin{aligned}
\mathbf{Feat}^\prime_{T_{i}} &=  \mathcal{F}(\mathcal{S}(T_{i-1},T_{i},T_{i+1})),
\end{aligned}
\end{equation}
where \( i \in \{1, 2, \dots, n\} \). This is followed by sampling with a stride of 2 and feature update, allowing for the aggregation of information from a broader temporal context:
\begin{equation}
\begin{aligned}
\mathbf{Feat}_{T_{2i}} =  &\overline{\mathcal{F}}(\overline{\mathcal{S}}(\mathbf{Feat}^\prime_{T_{2i-1}}, \mathbf{Feat}_{T_{2i}}, \mathbf{Feat}^\prime_{T_{2i+1}})), \\
= &\overline{\mathcal{F}}(\overline{\mathcal{S}}(\mathcal{F}(\mathcal{S}(T_{2i-2},T_{2i-1},T_{2i})), \\
\quad &\mathcal{F}(\mathcal{S}(T_{2i-1},T_{2i},T_{2i+1})), \\
\quad &\mathcal{F}(\mathcal{S}(T_{2i},T_{2i+1},T_{2i+2})))),
\end{aligned}
\end{equation}\label{eq:frame}where \( i \in \{1, 2, \dots, n/2\} \). In this process, non-anchor frames are utilized multiple times, ensuring that the receptive field of each frame expands progressively. For instance, each anchor frame now accesses features from \( \{T_{2n-1}, T_{2n+1}\} \) rather than \( \{T_{2n}\} \), and each subsequent frame \( T_{2n} \) can leverage information from all preceding frames \( \{T_1, T_2, ..., T_{2n-1}\} \). The richer temporal interactions and more comprehensive information flow improve the model's ability to capture sequential features and long-term dependencies, thus enhancing performance in tasks requiring extensive temporal context.

\subsection{\textbf{Spatio-Temporal Selection Scanning}}\label{sec:Hilbert }
\label{sec:3.1.1}
Imposing a degree of order on the point cloud is essential to achieve a structured representation. Traditional methods typically employ a two-step approach: spatial sorting followed by temporal sorting. However, some points that are spatio-temporally distant may actually be the same point and exhibit higher similarity, but temporally sequential scanning separates semantically similar points across frames. As shown in \cref{fig:fig1}, points from \( t_1 \) and \( t_3 \) are farther apart than random points after temporally sequential scanning. This leads to the issue of long-range attenuation of similar points in the unidirectional modeling of SSM.

In contrast to the conventional scanning, we point out STSS, which utilizes a prompt network to assign similarity-based categories to points, clustering those that are spatio-temporally distant but with high semantic similarity. A lightweight prompt network, such as PointNet, is used to generate a prompt matrix $\mathcal{P}$ with $k$ prompt categories $(p_1, p_2, p_3, ... , p_k)$, clustering points by semantic similarity:
\begin{equation}
\begin{aligned}
    \mathcal{P} &= \text{PromptNetwork}(\mathbf{Feat}) \in \mathbb{R}^{N \times K} ,  \\
    \mathcal{P} &= \{p_1, p_2, p_3, ... , p_k\}.\\
\end{aligned}
\end{equation}

STSS then clusters points that share the same prompt category and sorts them separately within each semantic-aware cluster. After clustering, points that are close in position exhibit higher correlation show higher correlation, whereas using standard spatial filling curves that traverse the 3D space along the $x$, $y$, and $z$ axes might result in adjacent points in the sequence having large spatial distances, leading to significant feature differences. Therefore, we use Intra-cluster Hilbert Sorting \cite{pstv3}. Within each cluster \( C_k \), points are sorted using the 3D Hilbert curve:
\begin{equation} 
\begin{aligned}
\quad \mathbf{X}_j &= \{x_i | \arg\max(p_i) = j\},\\
\overline{\mathbf{X}}_j &= \text{HilbertSort}(\mathbf{X}_j).
\end{aligned}
\end{equation}

To maintain motion continuity, we organize points across frames in sequential order. As illustrated in the  \cref{fig:fig1} and \cref{fig:framework}, we use Inter-cluster Chronological Ordering for temporal modeling within each cluster, which preserves the natural motion sequence (indicated by longitudinal or oblique arrows). Points within each frame are marked by the same color (\eg, red, yellow, green). Given an input point cloud \(\mathbf{X} \in \mathbb{R}^{\mathrm{T} \times \mathrm{N} \times 3}\) with features \(\mathbf{Feat} \in \mathbb{R}^{\mathrm{T} \times \mathrm{N} \times \mathrm{C}}\), after processing with STSS, we obtain \(\overline{\mathbf{X}} \in \mathbb{R}^{\mathrm{L} \times 3}\) and \(\overline{\mathbf{Feat}} \in \mathbb{R}^{\mathrm{L} \times \mathrm{C}}\), where \(\mathrm{L} = \mathrm{T \times N}\), the overall process of STSS is formulated as follows::
\begin{equation}
\begin{aligned}
    \mathcal{P} &= \text{PromptNetwork}(\mathbf{Feat}) \in \mathbb{R}^{N \times K} ,  \\
    \quad \mathbf{X}_j,\mathbf{Feat}_j &= \{x_i, feat_i | \arg\max(p_i) = j\},\\
    \widehat{\mathbf{X}_j}, \widehat{\mathbf{Feat}_j} &= \text{HilbertSort}(\mathbf{X}_j, \mathbf{Feat}_j), \\
    \overline{\mathbf{X}_j}, \overline{\mathbf{Feat}_j} &=\text{ChronologicalOrdering}(\widehat{\mathbf{X}_j}, \widehat{\mathbf{Feat}_j}),\\
    \overline{\mathbf{X}} &= \bigcup_{j=1}^{k}\overline{\mathbf{X}_j}, \overline{\mathbf{Feat}} = \bigcup_{j=1}^{k} \overline{\mathbf{Feat}_j}.\\
\end{aligned}
\end{equation}

\subsection{\textbf{Spatio-Temporal Structure Aggregation}}
\label{sec:3.1.2}
While STSS unifies spatio-temporal disorder, the unidirectional scanning inherent to SSMs inevitably disrupts local geometric relationships critical for fine-grained motion analysis. Conventional SSM-based methods \cite{mamba4d} serialize points without preserving neighborhood contexts, sacrificing spatial coherence for computational efficiency. To overcome this limitation, we introduce Spatio-Temporal Structure Aggregation (STSA), a lightweight yet effective module that explicitly recovers geometric details while capturing dynamic motion patterns.

As illustrated in \cref{fig:framework}, STSA operates within the ST-SSM block alongside bi-SSM, forming a symbiotic relationship: STSA concentrates on localized spatio-temporal interactions while SSMs handle global sequential dependencies. This division of labor ensures efficient modeling of both neighborhood structures and long-range dynamics. The STSA process comprises two stages:
\vspace{-4mm}
\paragraph{Neighborhood Construction via 4D KNN:}
Traditional spatial KNN fails to account for temporal continuity, treating each frame as isolated. Unlike previous Spatio-Temporal Structure Encoding, STSA places all points across frames into a unified coordinate space ($L=T \times N$) and extends KNN to 4D by incorporating temporal embeddings \(\mathbf{E}^t\) into the distance metric for neighbor selection. This not only enhances filtering efficiency but also further enables unified spatio-temporal modeling:
\begin{equation} K = \underset{\mathbf{X}_K \in X}{\text{argmin}} \left( |\mathbf{X}_C - \mathbf{X}_K| + |\mathbf{E}_C^t - \mathbf{E}_K^t| \right). 
\end{equation} 
This constructs neighborhoods that preserve both geometric cohesion (through spatial coordinates $\mathbf{X}$) and motion trajectories (via learned temporal embeddings $\mathbf{E}^t$).
\vspace{-4mm}
\paragraph{Spatio-temporal Feature Propagation:} For each center point $C$, we normalize neighbor features $\mathbf{F}_K$ relative to the center while retaining absolute positional context: \begin{equation} \mathbf{F}_K' = \frac{\mathbf{F}_K - \mathbf{F}_C}{|\mathbf{F}_K - \mathbf{F}_C|_2 + \epsilon } \oplus \mathbf{F}_C \in \mathbb{R}^{\mathrm{L} \times k \times 2C}, \end{equation} where $\oplus$ denotes concatenation. This dual representation captures both relative geometric variations and absolute feature characteristics, enabling nuanced local interactions. To dynamically prioritize informative neighbors, we replace conventional spatio-temporal convolution with attention-inspired exponential pooling:
\begin{equation} 
\mathbf{F}_C' = \text{MLP}\left( \sum_{i=1}^K \frac{e^{\mathbf{F}_K'^i}}{\sum_j e^{\mathbf{F}_K'^j}} \cdot \mathbf{F}_K'^i \right)\in \mathbb{R}^{\mathrm{L} \times k \times C}.
\end{equation} This differentiable operation amplifies salient features while adaptively focusing on kinetically significant points without additional parameters.

STSA's design directly addresses the geometric information loss caused by serialization in three key ways: (1) The 4D KNN explicitly reconstructs spatio-temporal neighborhoods disrupted during scanning. (2) Relative feature normalization preserves displacement-sensitive motion cues. (3) Adaptive pooling maintains efficiency. By interlacing STSA with SSM layers, our model alternately focuses on micro-interactions and macro-dynamics, achieving comprehensive 4D understanding with linear complexity.
\begin{table}[!ht]
\centering
\caption{Action recognition accuracy on MSR-Action3D dataset.}
\renewcommand{\arraystretch}{1.1}
\footnotesize
\begin{tabular*}{\columnwidth}{c|c|c@{\extracolsep{\fill}}c@{\extracolsep{\fill}}c@{\extracolsep{\fill}}}
        \toprule
        Backbone  & Method                   & Venue              & Frames  & Acc (\%) \\
        \cline{1-5}
         \multirow{11}{*}{CNN}  &\multirow{2}{*}{MeteorNet \citep{MeteorNet}} &\multirow{2}{*}{ICCV'19}        & 16                & 88.21 \\
                       &           &                       & 24                & 88.50 \\
        \cline{2-5}
        & \multirow{3}{*}{PSTNet \citep{p4trans}}&\multirow{3}{*}{ICLR'21}   & 16                & 89.90 \\
              &     &                       & 24                & 91.20 \\                               &     &                       & 36                & 88.21 \\      
        \cline{2-5} 
        &\multirow{2}{*}{Kinet \citep{kinet}}& \multirow{2}{*}{CVPR'22}                      & 16                & 91.92 \\
       &                                &                       & 24                & 93.27\\
       \cline{2-5}
        &\multirow{2}{*}{PointCMP \citep{PointCMP}}&\multirow{2}{*}{ICCV'23} & 16                & 92.26 \\
         &                                          &                       & 24                & 93.27 \\
        \cline{2-5} 
        &\multirow{2}{*}{3DinAction \citep{3DInAction}}&\multirow{2}{*}{CVPR'24}         & 16                & 90.57 \\                       
        &                          &                       & 24                 &92.23 \\
        \cline{1-5}
        \multirow{12}{*}{Transformer}  & \multirow{3}{*}{P4Transformer \citep{p4trans}}&\multirow{3}{*}{CVPR'21}   & 16                & 89.56 \\
              &     &                       & 24                & 90.94 \\     
                                                &             &          & 36                & 82.81 \\                                               
        \cline{2-5} 
        &\multirow{2}{*}{PPTr \citep{pptr}} & \multirow{2}{*}{ECCV'22}  & 16                & 90.31 \\
             &                         &                       & 24                & 92.33 \\
        \cline{2-5}    
          & \multirow{3}{*}{PST-Transformer \citep{psttrans}} & \multirow{3}{*}{TPAMI’23}        & 16                & 91.98 \\
       &                     &                       & 24                & 93.73 \\
                                                   &                  &     & 36                & 91.15 \\
        \cline{2-5}  
        &\multirow{2}{*}{PointCPSC \citep{PointCP}}&\multirow{2}{*}{ICCV'23 } & 16                & 92.26 \\
        &                               &                       & 24                & 92.68 \\
        \cline{2-5}
        &\multirow{2}{*}{LeaF \citep{leaf}} & \multirow{2}{*}{ICCV'23}   & 16                & 91.50 \\
             &                         &                       & 24                & 93.84 \\
        \cline{1-5}
        \multirow{5}{*}{SSM}    &   \multirow{2}{*}{MAMBA4D \citep{mamba4d}} &  \multirow{2}{*}{CVPR'25}   & 24                & 92.68 \\
      &                                              &                       & 36                & 93.23 \\
        \cline{2-5}
        &\multirow{3}{*}{UST-SSM (Ours)} &\multirow{3}{*}{-}          & 16                & \textbf{93.27} \\
                                                             &                  &     & 24                & \textbf{94.77} \\
                                                             &               &        & 36                & \textbf{95.12} \\
        \bottomrule
    \end{tabular*}
\renewcommand{\arraystretch}{1}
\label{tab:msr}
\vspace{-2mm}
\end{table}

\section{Experiment}
\begin{table}[htb]
\centering
\caption{Action recognition accuracy of various methods on NTU RGB+D dataset. CS and CV represent the cross-subject and cross-view evaluation protocols.}
\renewcommand{\arraystretch}{1}
\footnotesize
\begin{tabular*}{\columnwidth}{c|c@{\extracolsep{\fill}}c@{\extracolsep{\fill}}c}
\toprule
Method& Venue & CS & CV  \\
\midrule[0.3mm]
AS-GCN \citep{AS-GCN}&CVPR'19& 86.8 &94.2\\
AGC-LSTM \citep{AGC}&CVPR'19 &89.2& 95.0\\
VA-fusion \citep{VA-fusion}&TPAMI'19& 89.4 &95.0\\
DGNN \citep{DGNN}&CVPR'19& 89.9 &96.1 \\
GCA-LSTM \citep{GCA} & TPAMI'20 & 74.4 &  82.8  \\ 
P4Transformer \citep{p4trans}&CVPR'21& 90.2 & 96.4  \\
PSTNet  \citep{pstnet} & ICLR'21& 90.5 & 96.5  \\
PointCPSC \citep{PointCP}& ICCV'23& 88.0 & -\\
PointCMP \citep{PointCMP} &CVPR'23& 88.5 & - \\
PST-Transformer  \citep{psttrans}&TPAMI'23 & 91.0 & 96.4 \\
3DinAction \citep{3DInAction} &CVPR'24 & 89.3 & -\\
\hline
\addlinespace[0.5em]
 UST-SSM (Ours)&- &\textbf{92.1} & \textbf{97.8} \\ 
\bottomrule
\end{tabular*}
\label{tab:ntu60}
 \vspace{-2mm}
\end{table}
\begin{table}[ht]
    \centering
    \caption{Efficiency and accuracy comparison on MSR-Action3D dataset with 36 frames.}
    \renewcommand{\arraystretch}{0.9}
    \footnotesize
\begin{tabular*}{\columnwidth}{c|c@{\extracolsep{\fill}}c@{\extracolsep{\fill}}c@{\extracolsep{\fill}}c@{\extracolsep{\fill}}}
    \toprule
    \multirow{2}{*}{Method}  & Params  & Memory &Times   & Accuracy \\ 
      & (M)  & (GB) &(H)  & (\%)  \\ 
    \midrule 
    PSTNet           & 8.259 &22.031 &8.526 &  88.21 \\ 
    P4Transformer    & 0.834 & 3.667 &3.799 &  82.81 \\ 
    PST-Transformer  & 0.289 & 4.072 & 2.995 & 91.15 \\ 
    UST-SSM          & 0.138 & 2.403 & 2.824  &95.12 \\ 
    \bottomrule
    \end{tabular*}
    \label{tab:EFFICIENCY}
    \vspace{-4mm}
\end{table}
\begin{figure}[htb]
	\centering 
	\begin{tabular}{c}		
		\includegraphics[width=8cm]{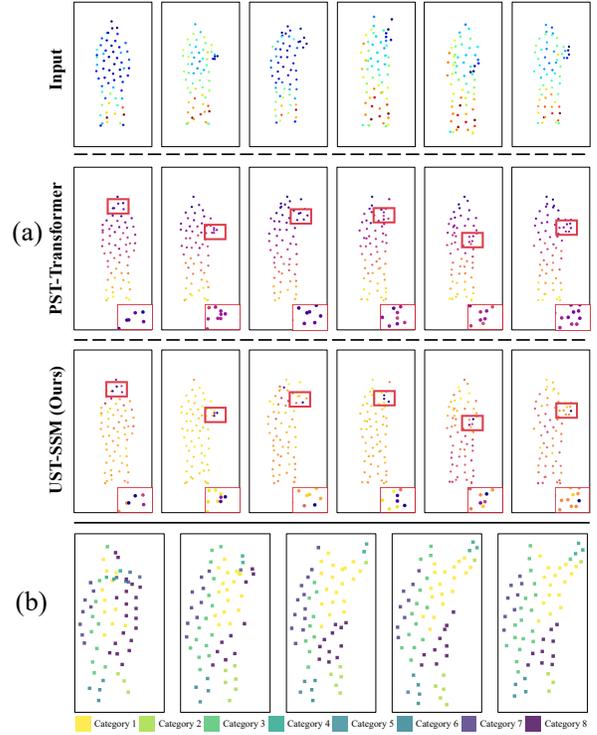}\\
	\end{tabular}
     \vspace{-4mm}
	\caption{\textbf{(a) Feature visualization:} For the inputs, colors represent depth, while darker colors indicate higher feature weights. Compared to attention mechanisms, our method produces higher activations in key moving regions, demonstrating its ability to capture more informative cues for action reasoning. \textbf{(b) Prompt matrix visualization:} Semantically similar regions across distant frames are clustered into the same prompt category.}\label{fig:visul}%
    \vspace{-4mm}
\end{figure}
\subsection{Experimental Details}
\textbf{Datasets:} The MSR-Action3D dataset \cite{msr} consists of 567 depth videos captured with Kinect v1, spanning 20 action categories and containing a total of 23K frames. We follow the protocol in P4Transformer \cite{p4trans} for training/testing split. The NTU RGB+D dataset \cite{ntu60} is a large-scale benchmark for action recognition, comprising 56,880 videos of 60 actions performed by 40 subjects across 80 camera views. For NTU RGB+D dataset, we adopt the cross-subject and cross-view evaluation protocols. Synthia 4D \cite{Minkowski} leverages the Synthia dataset to generate 3D video sequences, covering six driving scenarios. Each scenario includes four stereo RGB-D images captured from the top of a moving vehicle. We use the same training/testing splits as \cite{pstnet,p4trans, psttrans}.\\
\textbf{Implementation Details:} For the MSR-Action3D and Synthia 4D datasets, we set the number of ST-SSM to 1 and train the model for 50 epochs on an NVIDIA RTX 4090 GPU using the SGD optimizer with the initial learning rate of 0.01, decayed by a factor of 0.1 at the 20th and 30th epochs. For the NTU RGB+D dataset, we increase the number of ST-SSM to 3 and train the model for 15 epochs. The Prompt Network is composed of a T-Net for geometric transformation and a PointNet architecture for feature extraction, with both parts initialized using Xavier uniform distribution. For 4D semantic segmentation, we establish our segmentation framework based on recent representative transformer backbone P4Transformer. Specifically, we replace the transformer with ST-SSM and use STSS for both sorting and reverse sorting.
\begin{table}[htb]
\centering
\caption{4D semantic segmentation results on Synthia 4D dataset. }
\vspace{-2mm}
\renewcommand{\arraystretch}{0.9}
\footnotesize
\begin{tabular*}{\columnwidth}{c|c@{\extracolsep{\fill}}c@{\extracolsep{\fill}}c@{\extracolsep{\fill}}c@{\extracolsep{\fill}}}
\toprule
Method & Venue  & Input & Frame   & mIoU (\%) \\
\midrule
3D MinkNet14 \cite{Minkowski}  &  CVPR'19  & voxel & 1    & 76.24 \\
4D MinkNet14 \cite{Minkowski} &  CVPR'19 & voxel & 3  & 77.46 \\
\midrule
MeteorNet-m \cite{MeteorNet} &ICCV'19& point& 2& 81.47 \\
MeteorNet-l \cite{MeteorNet} &ICCV'19& point & 3     & 81.80 \\
PSTNet \cite{pstnet}&ICLR'21&point&3 & 82.24\\
P4Transformer \cite{p4trans}    &CVPR'21& point & 1& 82.41 \\
P4Transformer \cite{p4trans}    &CVPR'21& point & 3  & 83.16 \\
PST-Transformer \cite{psttrans}    &TPAMI'22& point & 1 & 82.92 \\
PST-Transformer \cite{psttrans}    &TPAMI'22& point & 3  & 83.95  \\
MAMBA4D \cite{mamba4d} &CVPR'25& point & 3   & 83.35 \\
\midrule
P4Transformer + Ours & - & point & 3 & \textbf{84.06} \\
\bottomrule
\end{tabular*}
\renewcommand{\arraystretch}{1}
\label{tab:syn}
\vspace{-2mm}
\end{table}
   
\begin{table}[!htb]
\centering
\caption{Impact of different scanning strategies and space-filling curves on MSR-Action3D dataset.}
\vspace{-2mm}
\renewcommand{\arraystretch}{0.8}
\footnotesize
\begin{tabular*}{\columnwidth}{@{\extracolsep{\fill}}c|c@{\extracolsep{\fill}}c@{\extracolsep{\fill}}}
        \toprule
        Scanning                   & Space-Filling  & Accuracy  \\
        \midrule
        \multirow{7}{*}{ Cross-temporal}     
                                             & XYZ                &85.02 \\
                                             & XZY                &83.97  \\
                                             & YXZ                &88.15  \\
                                             & YZX                &87.46  \\
                                             & ZXY                &85.02  \\
                                             & ZYX                &86.41  \\
                                             & Hilbert Curve      &89.20  \\
        \midrule 
        \multirow{7}{*}{Temporally Sequential}     
                                             & XYZ                &93.03 \\
                                             & XZY                &92.68  \\
                                             & YXZ                &93.38  \\
                                             & YZX                &91.99  \\
                                             & ZXY                &92.68  \\
                                             & ZYX                &93.38  \\
                                             & Hilbert Curve      &94.08\\
                                                     \midrule 
        \multirow{1}{*} {Spatio-Temporal Selection}    
                                            & Hilbert Curve      &\textbf{94.77}\\
        \bottomrule
    \end{tabular*}
\renewcommand{\arraystretch}{1}
\label{tab:cts}
\vspace{-2mm}
\end{table}
\begin{table}[ht]
    \centering
    \caption{The robustness of key components on recent representative Transformer architecture on MSR-Action3D dataset.}
    \vspace{-2mm}
    \footnotesize
    \renewcommand{\arraystretch}{1.1}
    \begin{tabular*}{\columnwidth}{c|ccccc}
        \toprule
        \multirow{2}{*}{Architecture} & \multicolumn{2}{c}{ST-SSM} & \multirow{2}{*}{STSS} & \multirow{2}{*}{TIS} & \multirow{2}{*}{Acc (\%)} \\
        \cline{2-3}
        & Selective SSM & STSA & & & \\
        \hline
        & \ding{55}  & \ding{55} & \ding{55} & \ding{55} & 90.94 \\
        \multirow{2}{*}{P4Transformer }&\ding{51} &\ding{55} & \ding{55}& \ding{55} & 92.33 \\
        \multirow{2}{*}{+ }&  \ding{51} & \ding{51} & \ding{55} &  \ding{55} & 93.03 \\
        \multirow{2}{*}{Ours}&\ding{51} &\ding{51}  & \ding{51}& \ding{55}& 94.43 \\\
        &  \ding{51}& \ding{51}& \ding{51}& \ding{51} & 94.77 \\
       \bottomrule
    \end{tabular*}
   \label{tab:P4TRANS}
    \vspace{-3mm}
\end{table}
\begin{table}[!ht]
\centering
\caption{Impact of different spatio-temporal sampling strategies on MSR-Action3D dataset.}
\vspace{-2mm}
\renewcommand{\arraystretch}{0.8}
\footnotesize
\begin{tabular*}{\columnwidth}{c|c|c@{\extracolsep{\fill}}c@{\extracolsep{\fill}}c@{\extracolsep{\fill}}}
        \toprule
       Strategy    &Stride &Params (M)& Memory (GB)& Accuracy (\%) \\
        \midrule
        \multirow{2}{*}{Single-stride}     
                                             & 1        &   0.046  & 3.079 &94.77 \\
                                             & 2        &   0.046 & 2.119 &94.43  \\
        \midrule 
        \multirow{1}{*} {TIS}    
                                             & 1+2  & 0.048&1.759  &\textbf{94.77}\\
        \bottomrule
    \end{tabular*}
\renewcommand{\arraystretch}{1}
\label{tab:TIS}
\vspace{-2mm}
\end{table}
\begin{table}[ht]
\caption{\raggedright Impact of STSS and STSA on NTU RGB+D dataset.}
\vspace{-2mm}
\renewcommand{\arraystretch}{0.8}
\footnotesize
\begin{tabular*}{\columnwidth}{c@{\extracolsep{\fill}}c@{\extracolsep{\fill}}c@{\extracolsep{\fill}}}
    \toprule
    Spatio-Temporal Selection Sorting  & STSA & Accuracy (\%) \\ 
    \midrule
   \ding{55} & \ding{55}  & 90.8 \\ 
   \ding{55} & \ding{51}  &   91.4 \\ 
   \ding{51} & \ding{55}  &   91.5 \\ 
   \ding{51} & \ding{51}  & \textbf{92.1} \\ 
    \bottomrule
    \end{tabular*}
    \label{tab:aba44}
    \vspace{-3mm}
\end{table}
\subsection{3D Action Recognition}\label{sec:3daction}
\textbf{Quantitative Comparison on MSR-Action3D dataset.} As shown in \cref{tab:msr}, we evaluate our model on the MSR-Action3D dataset and compare it against CNN-based methods \cite{pstnet,MeteorNet}, Transformer-based methods \cite{p4trans,psttrans,pptr} and SSM-based method \cite{mamba4d}. Our approach achieves higher recognition accuracy than state-of-the-art methods across commonly used 16-frame, 24-frame, and longer 36-frame inputs. Notably, Transformer-based methods experience a substantial performance drop as sequence length increases, underscoring the challenges in modeling long-sequence 4D videos. In contrast, our method demonstrates a steady improvement in recognition accuracy with longer sequences.\\
\textbf{Quantitative Comparison on NTU RGB+D dataset.} As shown in \cref{tab:ntu60}, UST-SSM achieves higher recognition accuracy than state-of-the-art methods across all evaluation settings. Notably, in the cross-subject evaluation on the NTU RGB+D dataset, UST-SSM surpasses the leading Transformer-based method, PST-Transformer, by 1.1\%.\\
\noindent\textbf{Qualitative Results.} As shown in \cref{fig:visul}, we compare the features obtained in ST-SSM with those derived from an attention mechanism. As expected, the Spatio-Temporal Structure Aggregation effectively focuses on the significant moving regions, supporting our intuition that ST-SSM can effectively capture the spatio-temporal structure of point cloud videos, serving as a viable alternative to attention mechanisms. As shown in the prompt matrix visualization in \cref{fig:visul}, similar regions across different frames are clustered into the same prompt category, validating the effectiveness of STSS in clustering similar points. \\
\textbf{Efficiency Comparison.} To validate the advantages of UST-SSM on long sequences, as shown in \cref{tab:EFFICIENCY}, with the setup of 36 frames, 2048 points, and a batch size of 32, UST-SSM significantly outperforms CNN-based and Transformer-based methods in terms of parameters, memory usage and training times on a single RTX 4090, while maintaining higher accuracy. As shown in \cref{fig:fig2}, as sequence length grows, GPU memory usage for P4Transformer and PST-Transformer follows a quadratic growth, whereas our method scales linearly.
\subsection{4D Semantic Segmentation}
\begin{figure*}[htbp]
	\centering 
	\begin{tabular}{c}		
		\includegraphics[width=17cm]{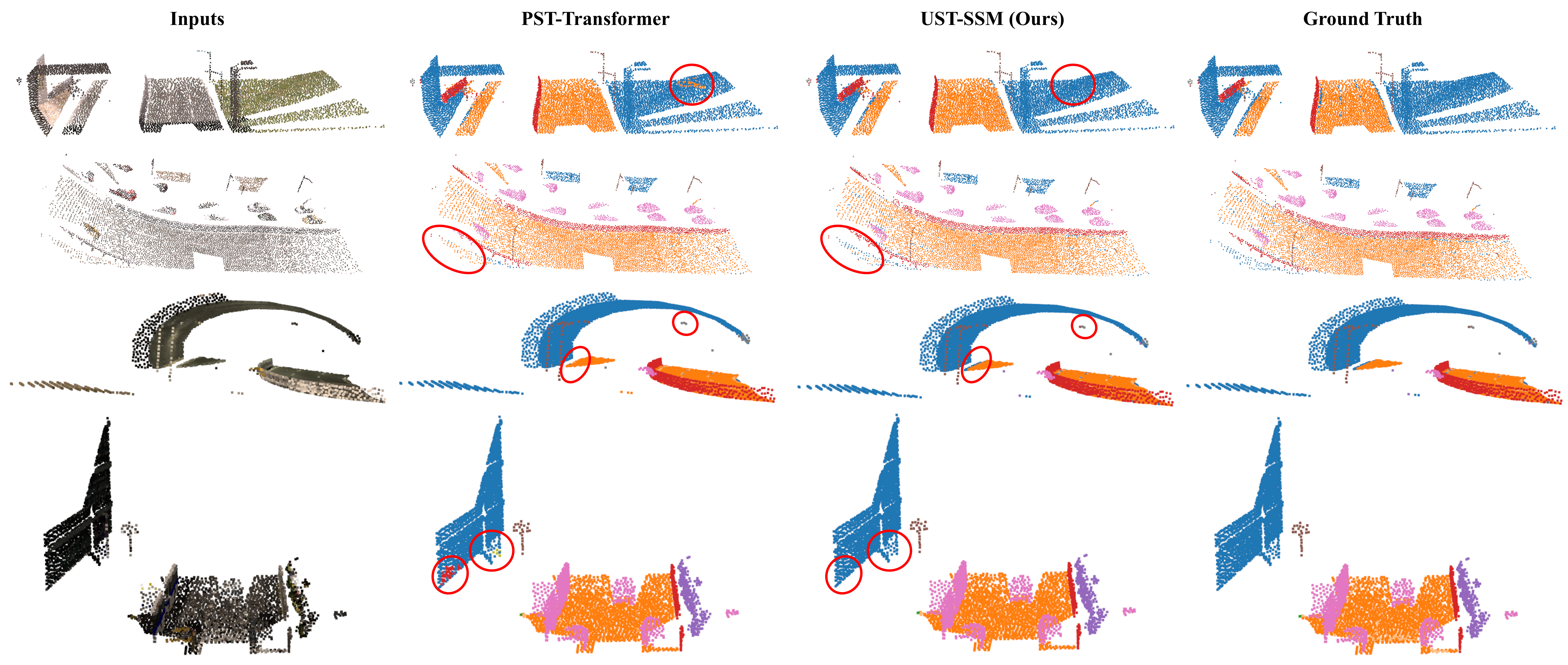}\\
	\end{tabular}
	\caption{4D semantic segmentation visualization on the Synthia 4D dataset. Our method achieves precise segmentation even in subtle regions, outperforming state-of-the-art Transformer-based methods.}\label{fig:syn1}%
    \vspace{-4mm}
\end{figure*}

\textbf{Comparison on Synthia 4D dataset.} To further demonstrate the versatility of UST-SSM for point cloud video modeling, we evaluate our method on 4D semantic segmentation. As shown in \cref{tab:syn}, our approach outperforms all CNN-based methods \cite{MeteorNet} and Transformer-based methods \cite{p4trans, psttrans, pptr}. We present four segmentation results from the Synthia 4D dataset in \cref{fig:syn1}. Our method achieves precise segmentation even in subtle regions, outperforming state-of-the-art Transformer-based methods.
\subsection{Ablation Study}
The STSS, STSA and TIS are critical components of our method. To highlight the importance of our customized strategies for extending point cloud videos in SSMs, we report the effects of various strategies used in UST-SSM in \cref{tab:P4TRANS}. The results show that the combination of STSS, STSA, and TIS contributes to improved action recognition accuracy, validating the effectiveness of unified spatio-temporal sorting and the impact of spatio-temporal feature interactions. Notably, the full UST-SSM yields a 3.83\% improvement in recognition accuracy, underscoring the importance of extending SSMs to point cloud videos.\\
\textbf{Evaluation of STSS:} As shown in \cref{tab:P4TRANS}, applying STSS results in a 1.4\% improvement in recognition accuracy. To further evaluate our STSS, we compare the performance of different scanning strategies, as presented in \cref{tab:cts}. Among these, the combination of the Hilbert space-filling curve and Spatio-Temporal Selection, adopted by UST-SSM, consistently outperforms all other configurations, demonstrating its superiority in unified spatio-temporal modeling.\\
\textbf{Evaluation of STSA and ST-SSM:} As shown in \cref{tab:P4TRANS}, adding STSA to the ST-SSM module results in a 0.7\% improvement in recognition accuracy. Additionally, using the full ST-SSM module achieves a 2.09\% improvement over P4Transformer. These results validate the effectiveness of STSA and its synergy with SSM.\\
\textbf{Evaluation of TIS:} 
We also conduct an ablation study on different spatio-temporal sampling strategies in \cref{tab:TIS}, evaluating the memory consumption introduced by TIS. Single-stride refers to the original sampling strategy in previous methods. Our TIS compensates for the temporal field of view of each frame, achieving superior results with fewer sampled frames while reducing GPU memory usage.\\
\textbf{Evaluation of the robustness of our designs:} Given that one of the key advantages of the selective state space models over Transformer-based models is its linear complexity, demonstrating its performance on large-scale datasets is essential to fully validate the scalability and practical benefits of the proposed method. Specifically, we evaluate the design on the NTU RGB+D dataset. The results shown in \cref{tab:aba44} highlight the effectiveness of our designs.\\
\textbf{Evaluation of Semantic Similarity:}
\begin{figure}[t]
    \centering
    \begin{tikzpicture}[font=\tiny]  
    \begin{axis}[
        xlabel={Cluster number $K$},
        xlabel style={yshift=0.5ex,font=\footnotesize},
        ylabel={Acc. (\%)},
        ylabel style={yshift=-0.5ex,font=\footnotesize},
        xmode=log,
        log basis x=2,
        xtick={1, 2, 4, 8, 16, 32, 64},
        xticklabels={$2^0$, $2^1$, $2^2$, $2^3$, $2^4$, $2^5$, $2^6$},
        ymin=93.5,    
        ymax=95,
        ytick={93.5, 94.0, 94.5, 95.0},
        yticklabel style={font=\tiny},
        xticklabel style={font=\tiny},
        ymajorgrids=true,
        xmajorgrids=true,
        grid style={dashed, gray!40},
        width=0.8\linewidth,  
        height=2.5cm,          
        axis line style={-},
        tick align=outside,
        tickwidth=0.3pt,
    ]

    \addplot [
        color=blue,
        mark=*,
        mark options={fill=blue, scale=0.6},
        thick
    ] coordinates {
        (1, 93.73)
        (2, 93.73)
        (4, 94.08)
        (8, 94.77)
        (16, 94.77)
        (32, 93.73)
        (64, 93.73)
    };
    \end{axis}
    \end{tikzpicture}
    \vspace{-3mm}
    \caption{Ablation study on semantic similarity.}
    \label{fig:k_sensitivity_curve}
    \vspace{-5mm}
\end{figure}
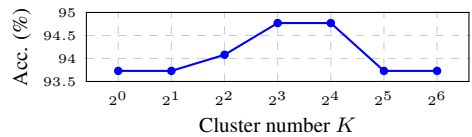
An ablation study on the weighting for semantic similarity is conducted in the Fig. \ref{fig:k_sensitivity_curve}. When $K=2^0$, excessively weighting spatial distance and ignoring semantic similarity reduces it to Temporally Sequential Scanning. When $K$ equals the number of points $2^6$, overemphasizing semantic similarity leads to semantic-only grouping, causing decreased performance. The best results are achieved when $K$ is set to $2^3$ or $2^4$.\\
\vspace{-4mm}
\section{Conclusion}
In this paper, we present UST-SSM, an SSM-based model for unified spatio-temporal encoding. UST-SSM clusters spatio-temporally distant but similar points through Spatio-Temporal Selection Scanning, enabling unified spatio-temporal sorting and long-sequence memory retention. Through Spatio-Temporal Structure Aggregation and Temporal Interaction Sampling, UST-SSM compensates for missing geometric and motion details while enhancing temporal interaction between frames. Extensive experiments validate the effectiveness of UST-SSM in point cloud video modeling. Our method transforms unordered point cloud videos into ordered sequences along the spatio-temporal dimension, effectively addressing the reliance of SSM models on structured point clouds and successfully extending SSMs to point cloud video modeling.

\noindent\textbf{Acknowledgements} This work was supported by National Natural Science Foundation of China (No. 62473007), Natural Science Foundation of Guangdong Province (No. 2024A1515012089), Shenzhen Innovation in Science and Technology Foundation for The Excellent Youth Scholars (No. RCYX20231211090248064).

{
    \small
    \bibliographystyle{ieeenat_fullname}
    \bibliography{main}

\begin{thebibliography}{42}
\providecommand{\natexlab}[1]{#1}
\providecommand{\url}[1]{\texttt{#1}}
\expandafter\ifx\csname urlstyle\endcsname\relax
  \providecommand{\doi}[1]{doi: #1}\else
  \providecommand{\doi}{doi: \begingroup \urlstyle{rm}\Url}\fi

\bibitem[Ben-Shabat et~al.(2024)Ben-Shabat, Shrout, and Gould]{3DInAction}
Yizhak Ben-Shabat, Oren Shrout, and Stephen Gould.
\newblock 3dinaction: Understanding human actions in 3d point clouds.
\newblock In \emph{Proceedings of the IEEE/CVF Conference on Computer Vision and Pattern Recognition (CVPR)}, 2024.

\bibitem[Chen et~al.(2023)Chen, Xia, Ichter, Rao, Gopalakrishnan, Ryoo, Stone, and Kappler]{104}
Boyuan Chen, Fei Xia, Brian Ichter, Kanishka Rao, Keerthana Gopalakrishnan, Michael~S. Ryoo, Austin Stone, and Daniel Kappler.
\newblock Open-vocabulary queryable scene representations for real world planning.
\newblock In \emph{2023 IEEE International Conference on Robotics and Automation (ICRA)}, 2023.

\bibitem[Chen et~al.(2024)Chen, Huang, Xu, Pei, Chen, Li, Wang, Li, Lu, and Wang]{videomambasuitestate}
Guo Chen, Yifei Huang, Jilan Xu, Baoqi Pei, Zhe Chen, Zhiqi Li, Jiahao Wang, Kunchang Li, Tong Lu, and Limin Wang.
\newblock Video mamba suite: State space model as a versatile alternative for video understanding, 2024.

\bibitem[Choy et~al.(2019)Choy, Gwak, and Savarese]{Minkowski}
Christopher Choy, JunYoung Gwak, and Silvio Savarese.
\newblock 4d spatio-temporal convnets: Minkowski convolutional neural networks.
\newblock In \emph{Proceedings of the IEEE/CVF Conference on Computer Vision and Pattern Recognition (CVPR)}, 2019.

\bibitem[Dai et~al.(2017)Dai, Chang, Savva, Halber, Funkhouser, and Niessner]{scannet}
Angela Dai, Angel~X. Chang, Manolis Savva, Maciej Halber, Thomas Funkhouser, and Matthias Niessner.
\newblock Scannet: Richly-annotated 3d reconstructions of indoor scenes.
\newblock In \emph{2017 IEEE Conference on Computer Vision and Pattern Recognition (CVPR)}, 2017.

\bibitem[Fan et~al.(2021)Fan, Yang, and Kankanhalli]{p4trans}
Hehe Fan, Yi Yang, and Mohan Kankanhalli.
\newblock Point 4d transformer networks for spatio-temporal modeling in point cloud videos.
\newblock In \emph{2021 IEEE/CVF Conference on Computer Vision and Pattern Recognition (CVPR)}, 2021.

\bibitem[Fan et~al.(2022{\natexlab{a}})Fan, Yu, Ding, Yang, and Kankanhalli]{pstnet}
Hehe Fan, Xin Yu, Yuhang Ding, Yi Yang, and MohanS. Kankanhalli.
\newblock Pstnet: Point spatio-temporal convolution on point cloud sequences.
\newblock \emph{Cornell University - arXiv,Cornell University - arXiv}, 2022{\natexlab{a}}.

\bibitem[Fan et~al.(2022{\natexlab{b}})Fan, Yu, Yang, and Kankanhalli]{Fan_Yu_Yang_Kankanhalli_2022}
Hehe Fan, Xin Yu, Yi Yang, and Mohan Kankanhalli.
\newblock Deep hierarchical representation of point cloud videos via spatio-temporal decomposition.
\newblock \emph{IEEE Transactions on Pattern Analysis and Machine Intelligence}, 2022{\natexlab{b}}.

\bibitem[Fan et~al.(2023)Fan, Yang, and Kankanhalli]{psttrans}
Hehe Fan, Yi Yang, and Mohan Kankanhalli.
\newblock Point spatio-temporal transformer networks for point cloud video modeling.
\newblock \emph{IEEE Transactions on Pattern Analysis and Machine Intelligence}, 2023.

\bibitem[Gu and Dao(2023)]{mamba}
Albert Gu and Tri Dao.
\newblock Mamba: Linear-time sequence modeling with selective state spaces.
\newblock 2023.

\bibitem[Gu et~al.(2020)Gu, Dao, Ermon, Rudra, and R\'{e}]{HIPPO}
Albert Gu, Tri Dao, Stefano Ermon, Atri Rudra, and Christopher R\'{e}.
\newblock Hippo: Recurrent memory with optimal polynomial projections.
\newblock In \emph{Advances in Neural Information Processing Systems}, 2020.

\bibitem[Gu et~al.(2021{\natexlab{a}})Gu, Goel, and R'e]{Gu2021EfficientlyML}
Albert Gu, Karan Goel, and Christopher R'e.
\newblock Efficiently modeling long sequences with structured state spaces.
\newblock \emph{ArXiv}, 2021{\natexlab{a}}.

\bibitem[Gu et~al.(2021{\natexlab{b}})Gu, Johnson, Goel, Saab, Dao, Rudra, and R\'{e}]{NEURIPS2021_05546b0e}
Albert Gu, Isys Johnson, Karan Goel, Khaled Saab, Tri Dao, Atri Rudra, and Christopher R\'{e}.
\newblock Combining recurrent, convolutional, and continuous-time models with linear state space layers.
\newblock In \emph{Advances in Neural Information Processing Systems}, 2021{\natexlab{b}}.

\bibitem[Guo et~al.(2024)Guo, Guo, Zha, Zhang, Li, Dai, Xia, and Li]{mambairv2}
Hang Guo, Yong Guo, Yaohua Zha, Yulun Zhang, Wenbo Li, Tao Dai, Shu-Tao Xia, and Yawei Li.
\newblock Mambairv2: Attentive state space restoration.
\newblock \emph{ArXiv}, 2024.

\bibitem[Han et~al.(2024)Han, Tang, Wang, and Li]{Mamba3D}
Xu Han, Yuan Tang, Zhaoxuan Wang, and Xianzhi Li.
\newblock Mamba3d: Enhancing local features for 3d point cloud analysis via state space model.
\newblock 2024.

\bibitem[Hu et~al.(2024)Hu, Baumann, Gui, Grebenkova, Ma, Fischer, and Ommer]{zigma}
Vincent~Tao Hu, Stefan~Andreas Baumann, Ming Gui, Olga Grebenkova, Pingchuan Ma, Johannes Fischer, and Björn Ommer.
\newblock Zigma: A dit-style zigzag mamba diffusion model, 2024.

\bibitem[Huang et~al.(2023)Huang, Mees, Zeng, and Burgard]{103}
Chenguang Huang, Oier Mees, Andy Zeng, and Wolfram Burgard.
\newblock Visual language maps for robot navigation.
\newblock In \emph{2023 IEEE International Conference on Robotics and Automation (ICRA)}, 2023.

\bibitem[Li et~al.(2024)Li, Li, Wang, He, Wang, Wang, and Qiao]{videomamba}
Kunchang Li, Xinhao Li, Yi Wang, Yinan He, Yali Wang, Limin Wang, and Yu Qiao.
\newblock Videomamba: State space model for efficient video understanding, 2024.

\bibitem[Li et~al.(2019)Li, Chen, Chen, Zhang, Wang, and Tian]{AS-GCN}
Maosen Li, Siheng Chen, Xu Chen, Ya Zhang, Yanfeng Wang, and Qi Tian.
\newblock Actional-structural graph convolutional networks for skeleton-based action recognition.
\newblock In \emph{2019 IEEE/CVF Conference on Computer Vision and Pattern Recognition (CVPR)}, 2019.

\bibitem[Li et~al.(2010)Li, Zhang, and Liu]{msr}
Wanqing Li, Zhengyou Zhang, and Zicheng Liu.
\newblock Action recognition based on a bag of 3d points.
\newblock In \emph{2010 IEEE Computer Society Conference on Computer Vision and Pattern Recognition - Workshops}, 2010.

\bibitem[Liang et~al.(2024)Liang, Zhou, Xu, Zhu, Ye, Tan, and Bai]{pointmamba}
Dingkang Liang, Xin Zhou, Wei Xu, Xingkui Zhu, Xiaoqing Ye, Xiao Tan, and Xiang Bai.
\newblock Pointmamba: A simple state space model for point cloud analysis.
\newblock 2024.

\bibitem[Liang and Boularias(2023)]{101}
Junchi Liang and Abdeslam Boularias.
\newblock Learning category-level manipulation tasks from point clouds with dynamic graph cnns.
\newblock In \emph{2023 IEEE International Conference on Robotics and Automation (ICRA)}, 2023.

\bibitem[Liu et~al.(2017)Liu, Wang, Hu, Duan, and Kot]{GCA}
Jun Liu, Gang Wang, Ping Hu, Ling-Yu Duan, and Alex~C. Kot.
\newblock Global context-aware attention lstm networks for 3d action recognition.
\newblock In \emph{2017 IEEE Conference on Computer Vision and Pattern Recognition (CVPR)}, 2017.

\bibitem[Liu et~al.(2023{\natexlab{a}})Liu, Wang, Jiang, Liu, and Wang]{105}
Jiuming Liu, Guangming Wang, Chaokang Jiang, Zhe Liu, and Hesheng Wang.
\newblock Translo: A window-based masked point transformer framework for large-scale lidar odometry.
\newblock \emph{Proceedings of the AAAI Conference on Artificial Intelligence}, 2023{\natexlab{a}}.

\bibitem[Liu et~al.(2024{\natexlab{a}})Liu, Han, Liu, Aviles-Rivero, Jiang, Liu, and Wang]{mamba4d}
Jiuming Liu, Jinru Han, Lihao Liu, Angelica~I. Aviles-Rivero, Chaokang Jiang, Zhe Liu, and Hesheng Wang.
\newblock Mamba4d: Efficient long-sequence point cloud video understanding with disentangled spatial-temporal state space models, 2024{\natexlab{a}}.

\bibitem[Liu et~al.(2019)Liu, Yan, and Bohg]{MeteorNet}
Xingyu Liu, Mengyuan Yan, and Jeannette Bohg.
\newblock Meteornet: Deep learning on dynamic 3d point cloud sequences.
\newblock In \emph{2019 IEEE/CVF International Conference on Computer Vision (ICCV)}, 2019.

\bibitem[Liu et~al.(2023{\natexlab{b}})Liu, Chen, Zhang, Huang, and Yi]{leaf}
Yunze Liu, Junyu Chen, Zekai Zhang, Jingwei Huang, and Li Yi.
\newblock Leaf: Learning frames for 4d point cloud sequence understanding.
\newblock In \emph{Proceedings of the IEEE/CVF International Conference on Computer Vision}, 2023{\natexlab{b}}.

\bibitem[Liu et~al.(2024{\natexlab{b}})Liu, Tian, Zhao, Yu, Xie, Wang, Ye, and Liu]{vmamba}
Yue Liu, Yunjie Tian, Yuzhong Zhao, Hongtian Yu, Lingxi Xie, Yaowei Wang, Qixiang Ye, and Yunfan Liu.
\newblock Vmamba: Visual state space model, 2024{\natexlab{b}}.

\bibitem[Luo et~al.(2018)Luo, Yang, and Urtasun]{Fast}
Wenjie Luo, Bin Yang, and Raquel Urtasun.
\newblock Fast and furious: Real time end-to-end 3d detection, tracking and motion forecasting with a single convolutional net.
\newblock In \emph{2018 IEEE/CVF Conference on Computer Vision and Pattern Recognition}, 2018.

\bibitem[Seichter et~al.(2021)Seichter, Köhler, Lewandowski, Wengefeld, and Gross]{102}
Daniel Seichter, Mona Köhler, Benjamin Lewandowski, Tim Wengefeld, and Horst-Michael Gross.
\newblock Efficient rgb-d semantic segmentation for indoor scene analysis.
\newblock In \emph{2021 IEEE International Conference on Robotics and Automation (ICRA)}, 2021.

\bibitem[Shahroudy et~al.(2016)Shahroudy, Li, Ng, and Wang]{ntu60}
Amir Shahroudy, Jun Li, Tian-Tsong Ng, and Gang Wang.
\newblock Ntu rgb+d: A large scale dataset for 3d human activity analysis.
\newblock 2016.

\bibitem[Shen et~al.(2023)Shen, Sheng, Wang, Guo, Liu, and Zhou]{PointCMP}
Zhiqiang Shen, Xiaoxiao Sheng, Longguang Wang, Yulan Guo, Qiong Liu, and Xi Zhou.
\newblock Pointcmp: Contrastive mask prediction for self-supervised learning on point cloud videos.
\newblock In \emph{2023 IEEE/CVF Conference on Computer Vision and Pattern Recognition (CVPR)}, 2023.

\bibitem[Sheng et~al.(2023)Sheng, Shen, Xiao, Wang, Guo, and Fan]{PointCP}
Xiaoxiao Sheng, Zhiqiang Shen, Gang Xiao, Longguang Wang, Yu~Kuen Guo, and Hehe Fan.
\newblock Point contrastive prediction with semantic clustering for self-supervised learning on point cloud videos.
\newblock \emph{2023 IEEE/CVF International Conference on Computer Vision (ICCV)}, 2023.

\bibitem[Shi et~al.(2019)Shi, Zhang, Cheng, and Lu]{DGNN}
Linjun Shi, Yifan Zhang, Jing Cheng, and Hanqing Lu.
\newblock Skeleton-based action recognition with directed graph neural networks.
\newblock \emph{IEEE Conference Proceedings,IEEE Conference Proceedings}, 2019.

\bibitem[Si et~al.(2019)Si, Chen, Wang, Wang, and Tan]{AGC}
Chenyang Si, Wentao Chen, Wei Wang, Liang Wang, and Tieniu Tan.
\newblock An attention enhanced graph convolutional lstm network for skeleton-based action recognition.
\newblock In \emph{2019 IEEE/CVF Conference on Computer Vision and Pattern Recognition (CVPR)}, 2019.

\bibitem[Teed and Deng(2021)]{106}
Zachary Teed and Jia Deng.
\newblock Droid-slam: Deep visual slam for monocular, stereo, and rgb-d cameras.
\newblock \emph{Neural Information Processing Systems,Neural Information Processing Systems}, 2021.

\bibitem[Wei et~al.(2022)Wei, Liu, Xie, Ke, and Guo]{Wei_2022_WACV}
Yimin Wei, Hao Liu, Tingting Xie, Qiuhong Ke, and Yulan Guo.
\newblock Spatial-temporal transformer for 3d point cloud sequences.
\newblock In \emph{Proceedings of the IEEE/CVF Winter Conference on Applications of Computer Vision (WACV)}, 2022.

\bibitem[Wen et~al.(2022)Wen, Liu, Huang, Duan, and Yi]{pptr}
Hao Wen, Yunze Liu, Jingwei Huang, Bo Duan, and Li Yi.
\newblock Point primitive transformer for long-term 4d point cloud video understanding.
\newblock 2022.

\bibitem[Wu et~al.(2024)Wu, Jiang, Wang, Liu, Liu, Qiao, Ouyang, He, and Zhao]{pstv3}
Xiaoyang Wu, Li Jiang, Peng-Shuai Wang, Zhijian Liu, Xihui Liu, Yu Qiao, Wanli Ouyang, Tong He, and Hengshuang Zhao.
\newblock Point transformer v3: Simpler, faster, stronger.
\newblock In \emph{2024 IEEE/CVF Conference on Computer Vision and Pattern Recognition (CVPR)}, 2024.

\bibitem[Zhang et~al.(2019)Zhang, Lan, Xing, Zeng, Xue, and Zheng]{VA-fusion}
Pengfei Zhang, Cuiling Lan, Junliang Xing, Wenjun Zeng, Jianru Xue, and Nanning Zheng.
\newblock View adaptive neural networks for high performance skeleton-based human action recognition.
\newblock \emph{IEEE Transactions on Pattern Analysis and Machine Intelligence}, 2019.

\bibitem[Zhong et~al.(2022)Zhong, Zhou, Hu, Wang, Trigoni, and Markham]{kinet}
Jianqi Zhong, Kaichen Zhou, Qingyong Hu, Bing Wang, Niki Trigoni, and A. Markham.
\newblock No pain, big gain: Classify dynamic point cloud sequences with static models by fitting feature-level space-time surfaces.
\newblock \emph{2022 IEEE/CVF Conference on Computer Vision and Pattern Recognition (CVPR)}, 2022.

\bibitem[Zhu et~al.(2024)Zhu, Liao, Zhang, Wang, Liu, and Wang]{visionmamba}
Lianghui Zhu, Bencheng Liao, Qian Zhang, Xinlong Wang, Wenyu Liu, and Xinggang Wang.
\newblock Vision mamba: Efficient visual representation learning with bidirectional state space model, 2024.

\end{thebibliography}
}

\end{document}